\documentclass{article}

\usepackage[margin=1in]{geometry}

\usepackage{hyperref}       
\usepackage{url}            
\usepackage{booktabs}       
\usepackage{amsfonts}       
\usepackage{nicefrac}       
\usepackage{microtype}      
\usepackage{amsmath}
\usepackage{booktabs}
\usepackage{multirow}
\usepackage{graphicx}
\usepackage{amssymb}
\usepackage{mathbbol}
\usepackage[numbers,sort&compress]{natbib}

\DeclareMathOperator{\sign}{sign}

\title{\sc Shapley Interpretation and Activation \\ in Neural Networks}

\author{Yadong Li and Xin Cui\thanks{Barclays, the views express in this article is the authors' own, they do not necessarily represent the views of Barclays. We thank Mohammad Jahangiri, Antonio Martini and Ariye Shater for many helpful comments and discussions, which made this paper significantly better. All remaining errors are authors' own.}}

\begin{document}

\maketitle

\begin{abstract}
We propose a novel Shapley value approach to help address neural networks' interpretability and ``vanishing gradient'' problems. Our method is based on an accurate analytical approximation to the Shapley value of a neuron with ReLU activation. This analytical approximation admits a linear propagation of relevance across neural network layers, resulting in a simple, fast and sensible interpretation of neural networks' decision making process.

We then derived a globally continuous and non-vanishing Shapley gradient, which can replace the conventional gradient in training neural network layers with ReLU activation, and leading to better training performance. We further derived a Shapley Activation (SA) function, which is a close approximation to ReLU but features the Shapley gradient. The SA is easy to implement in existing machine learning frameworks. Numerical tests show that SA consistently outperforms ReLU in training convergence, accuracy and stability.

\end{abstract}

\section{Introduction}
Deep neural network has achieved phenomenal success during the past decade. However, many challenges remain, among which interpretability and the ``vanishing gradient'' are two of the most well-known and critical problems \cite{vanishing, interpret}.

It is difficult to interpret how a neural network arrives at its output, especially for deep networks with many layers and an enormous number of parameters. Often times, users view a neural network as a ``black box'' that magically works. Lack of interpretability is often one of the top concerns \cite{baehrens2010explain, vellido2012making, arras2016explaining, ribeiro2016should, lipton2016mythos, lundberg2017unified, kindermans2017learning, arras2017explaining, samek2017explainable, zhang2018interpretable} among users who have to make real world decisions based on neural networks' outputs.

When a neural network grows deep and large, many of its neurons inevitably become dormant and stopped contributing to the final output, causing the problem of ``vanishing gradient''. As a result, individual training data may only update a small fraction of the neurons, thus prolonging the training process. The vanishing gradient may also cause an optimizer to stop prematurely. Therefore, large neural networks usually require multiple repetitions of training runs, to increase the chance of converging to a good solution.

There are several popular techniques for mitigating the problem of vanishing gradient \cite{bengio1994learning, hochreiter1998vanishing, pascanu2012understanding, pascanu2013difficulty, jozefowicz2015empirical, clevert2015fast, hu2018overcoming, kong2017hexpo, hanin2018neural}. One of them is the ReLU activation function \cite{relu} and its variants, such as leaky ReLU \cite{leakyrelu}, noisy ReLU \cite{noisyrelu} and SeLU \cite{selu}. The ReLU family of activation functions are known for preserving gradients better over deep networks than other types of activation functions. Preserving gradient is also the flagship features in many popular neural network architectures, such as ResNet \cite{resnet}. Batch normalization \cite{bn} is a recent breakthrough that is proven effective for preserving gradient in deep networks. Despite these mitigations, the vanishing gradient remains a challenge and continues to impose a practical limit on the size of a neural network, beyond which the training becomes infeasible.

In this paper, we propose a Shapley value \cite{shapley} approach for better interpreting a neural networks' prediction and improving its training by preserving gradient. Shapley value is a well established method in the cooperative game theory, and it forms a solid theoretical foundation to analyze and address both problems. Due to the great success and ubiquity of the ReLU activation function, we focus our efforts in finding a solution that can be viewed as a variant or approximation to the ReLU activation.

\section{Methodology}

In a typical neural network, a neuron with $n$ inputs can be written as: $y = f(\sum_{i=1}^n w_i x_i + b)$, where $f(\cdot)$ is a nonlinear activation function, $\vec x$ are the input to the neuron and $\vec w$, $b$ are the internal parameters of the neuron.

\subsection{Shapley Value Interpretation}

Shapley value was originally developed for fairly distributing shared gain or loss of a team among its team members. It has since been widely used in many different fields, such as the allocation of shared sale proceeds of package deals among participating service providers.

If we are interested in the contribution (a.k.a. relevance) of a given input $x_k$ to the output $y$ of a neuron, Shapley value is the only correct answer in theory \cite{shapley}. Shapley value of an input $x_k$ is defined to be the average of its incremental contributions to the output $y$ over all possible permutations of $x_i$s. We use $\alpha_k$ to denote the Shapley value of a neuron's input $x_k$, and Shapley value is conservative by construction: $y = \sum_{i=1}^{n} \alpha_i$.

\begin{table}
\caption {Shapley Value Example with Three Inputs and ReLU Activation\label{x3}}
\vspace{.5cm}

\centering

\begin{tabular}{@{}c|c|c|rrr|@{}}
\toprule
\multicolumn{1}{|c|}{\multirow{2}{*}{Permutation}} & $s = \sum_{i \in S} w_i x_i + b$ & \multirow{2}{*}{$\max(s, 0)$} & \multicolumn{3}{c|}{Incremental Contribution} \\ \cmidrule(lr){2-2} \cmidrule(l){4-6}
\multicolumn{1}{|c|}{} & $\vec w = (1, 2, 3), b = -1$ &  & \multicolumn{1}{c|}{$x_1$ = -1} & \multicolumn{1}{c|}{$x_2$ = 2} & \multicolumn{1}{c|}{$x_3$ = -1} \\ \midrule
\multicolumn{1}{|c|}{$x_1$, $x_2$, $x_3$} & -2, 2, -1 & 0, 2, 0 & 0 & 2 & -2 \\
\multicolumn{1}{|c|}{$x_1$, $x_3$, $x_2$} & -2, -5, -1 & 0, 0, 0 & 0 & 0 & 0 \\
\multicolumn{1}{|c|}{$x_2$, $x_1$, $x_3$} & 3, 2, -1 & 3, 2, 0 & -1 & 3 & -2 \\
\multicolumn{1}{|c|}{$x_2$, $x_3$, $x_1$} & 3, 0, -1 & 3, 0, 0 & 0 & 3 & -3 \\
\multicolumn{1}{|c|}{$x_3$, $x_1$, $x_2$} & -4, -5, -1 & 0, 0, 0 & 0 & 0 & 0 \\
\multicolumn{1}{|c|}{$x_3$, $x_2$, $x_1$} & -4, 0, -1 & 0, 0, 0 & 0 & 0 & 0 \\ \midrule
\multicolumn{1}{l|}{} & \multicolumn{2}{c|}{Average} & \multicolumn{1}{c}{$\alpha_1 = -\frac{1}{6}$} & \multicolumn{1}{c}{$\alpha_2 = \frac{4}{3}$} & \multicolumn{1}{c|}{$\alpha_3 = -\frac{7}{6}$} \\ \cmidrule(l){2-6}
\end{tabular}%

\vspace{.5cm}
\end{table}

Table \ref{x3} is an example of computing Shapley value of a neuron with 3 inputs. The column ``$s = \sum_{i=1}^n(w_i x_i + b)$'' and ``$\max(s, 0)$'' are the inputs and outputs of the ReLU when one, two or three inputs are activated in the order specified in the column ``Permutation''. In this example, there are 6 possible permutations for 3 inputs, thus the Shapley value for any of the inputs is the average of its incremental contributions over all six possible permutations. It is important to observe that the Shapley value of all 3 inputs are nonzero, even though the ReLU is currently deactivated with an overall output of 0. The reason for non-zero Shapley values is that the neuron could be activated by two combinations of inputs ($x_2$) or $(x_1, x_2)$ in this example.

There is some similarity between the computation of Shapley value and the random dropout neural network \cite{dropout}, in the sense that a random portion of the inputs are removed. However, it is worth pointing out their difference: a random dropout neural network turns off inputs randomly with an independent probability $p$, which leads to much higher chance of having roughly $(1-p)n$ active inputs. In contrast, random permutation works in two steps: first a single uniform random integer $\tilde h$ between 0 and $n$ is drawn; then a random ordering of the $n$ inputs is draw and only the first $\tilde h$ inputs in the permutation are kept on. Random permutation therefore gives equal probability in activating any number of inputs between 0 and $n$, yielding better chance of activating the neuron than random dropouts.

For a generic activation function $f(\cdot)$, Shapley value can only be evaluated numerically, for example using Monte Carlo simulation. Such a numerical implementation is computationally expensive and not conducive to analysis. Recently, an accurate analytical approximation to the Shapley value of the gain/loss function in the form of $\max(\sum_i a_i, \sum_i b_i)$ was discovered and verified in \cite{red}, the same approach can be adapted to ReLU activation function of $y = \max(\sum_{i=1}^n w_i x_i + b, 0)$, resulting in an analytical approximation of:
\begin{align}\nonumber
\mu &= \frac{1}{2} \sum_{i=1}^n w_i x_i \\
\label{app}
\sigma^2 &= \frac{1}{6}\sum_{i=1}^n (w_i x_i)^2 + \frac{1}{12} \left(\sum_{i=1}^n w_i x_i \right)^2 \\
\alpha_k &\approx \Phi(\frac{\mu + b}{\sigma}) \left(w_k x_k + \frac{b}{n}\right) \nonumber
\end{align}
where $\Phi(\cdot)$ is the standard normal distribution function, and $\alpha_k$ is the Shapley value of the k-th input.

The relevance of a neuron's output $y$ is defined to be its contribution to the final output of the entire neural network, which is typically a prediction or probability (e.g. in classification problems). If we denote relevance of a neuron's output as $r(y)$, a simple method to propagate the relevance to the neuron's inputs $x_k$ is to take advantage of the linearity of Shapley value and multiply a factor $\frac{r(y)}{y}$ to both side of $y = \sum_{i=1}^n\alpha_i$; and we arrive at the following propagation formula after some rearrangement:
\begin{equation}
\label{lrp}
r(x_k \leftarrow y) = \frac{\alpha_k}{\sum_{i=1}^n \alpha_i} r(y) \approx \frac{w_k x_k + \frac{b}{n}}{\sum_{i=1}^n w_i x_i + b} r(y)
\end{equation}
where $r(x_k \leftarrow y)$ is the relevance propagation from the neuron's output $y$ to input $x_k$.  Total relevance is conserved between layers as $r(x_k)$ is the sum of $r(x_k \leftarrow y)$ from all the connected neurons in the following layer. Since the $\Phi(\cdot)$ factor cancels, the layer-wise relevance propagation (LRP) \cite{lrp0, lrp} formula is identical for linear and linear + ReLU layers. If $r(y)$ is initialized to be the Shapley value of the neuron's output $y$, then the $r(x_k)$ retains the interpretation of being the (approximated) Shapley value of neuron's input $x_k$ after applying the LRP in \eqref{lrp}. A $\epsilon \cdot \sign(\sum_{i=1}^n \alpha_i)$ term with a small $\epsilon > 0$ can be added to the  denominator of \eqref{lrp} to prevent it from vanishing, similar to the $\epsilon$-variant formula of \cite{lrp0, lrp}.

The output layer of a neural network is often a nonlinear function, such as the softmax for classification. Before we can start the LRP via \eqref{lrp}, the relevance of the output layer has to be initialized to its Shapley value, which requires numerical evaluation (such as Monte Carlo) for most output functions, with few notable exceptions such as Linear and Linear+ReLU output layers. The Shapley value of a Linear+ReLU output layer is given by \eqref{app}.

An implicit assumption behind the LRP formula \eqref{lrp} is that the neuron's activation are independent from each other\footnote{An example of the effect of correlation is to consider two layered neurons that can never activate together, then there should be no relevance propagation through them, but formula \eqref{lrp} does.}, which generally does not hold across neural network layers. Therefore the Shapley values computed from LRP \eqref{lrp} is only a crude approximation for deep neural networks. A Monte Carlo simulation is required to compute the exact Shapley values of a neural network's input. However, the LRP \eqref{lrp} has the advantages of being very fast and producing the (approximated) relevance of all hidden layers as well as the input layer in one shot; while a MC approach would require a separate simulation for each neural network layer. In practice, a crude approximation like \eqref{lrp} may often be good enough to give users the intuition and confidence in using neural network's output.

The LRP formula \eqref{lrp} is similar to the native LRP algorithms given in \cite{lrp0, lrp}, but it replaces  $w_k x_k$ by $(w_k x_k + \frac{b}{n})$. The benefit of such a replacement is rather intuitive by considering the limiting case of $b \gg w_i x_i$ for all $i$, in which case individual $w_i x_i$ no longer makes much difference to the output, thus all inputs' relevance propagation should be approximately equal. By including the $\frac{b}{n}$, \eqref{lrp} produces more sensible results for this limiting case than the known LRP formulae in the literature.

The approximation \eqref{app} offers a straight forward explanation on why the same propagation formula applies to both Linear and Linear+ReLU layers, which is a common feature in existing LRP algorithms.  More generic approximations to Shapley values have been developed in \cite{lundberg2017unified, polysha} for interpreting neural networks, in comparison the analytical approximation in \eqref{app} is faster and more convenient for the ReLU layers.

\subsection{Shapley Gradient}

As shown in Table \ref{x3}, Shapley values are non-zeros for a neuron with ReLU as long as at least one of the input combinations can activate the neuron, it is much more likely than the neuron being active, which requires a much stronger condition of $s = \sum_{i=1}^n(w_i x_i + b) > 0$. This observation motivated the following approach to prevent the neuron's gradients from vanishing: we use $\frac{\partial \alpha_k}{\partial x_k}$ to replace the true gradient of $\frac{\partial y}{\partial x_k}$ during the back propagation stage of the training.  The result of this replacement is similar to a training procedure using random permutations, as mentioned earlier, random permutation is quite different from typical random dropouts.

In mathematical terms, this alternative gradient is an approximation of:
\begin{equation}
\label{xxx}
\frac{\partial y}{\partial \vec x} = \frac{\partial y}{\partial \vec \alpha} \frac{\partial \vec \alpha}{\partial \vec x} \approx \frac{\partial y}{\partial \vec \alpha} \left(\frac{\partial \vec \alpha}{\partial \vec x} \circ I \right) = \left[\frac{\partial \alpha_i}{\partial x_i}\right]
\end{equation}
$\frac{\partial \vec \alpha}{\partial \vec x}$ is the full Jacobian matrix and $\frac{\partial \vec \alpha}{\partial \vec x} \circ I $ is a matrix with only the diagonal elements of $\frac{\partial \vec \alpha}{\partial \vec x}$, the $\circ$ is element wise matrix product. The last step is because $y = \sum_{i=1}^n \alpha_i$ by construction, thus $\frac{\partial y}{\partial \vec \alpha}$ is a vector of 1s. Similar approximations are also applied to $\vec w$ and $b$ for back propagation:
\begin{align}
\label{backpop}
\frac{\partial y}{\partial x_k} &\approx \frac{\partial \alpha_k}{\partial x_k}  = \Phi(\frac{\mu+b}{\sigma}) w_k + g_kw_k\left(\frac{1}{2\sigma} - \frac{(\mu + b) (w_kx_k + \mu)}{6\sigma^3}\right) \nonumber \\
\frac{\partial y}{\partial w_k} &\approx \frac{\partial \alpha_k}{\partial w_k} = \Phi(\frac{\mu+b}{\sigma}) x_k +  g_k x_k\left(\frac{1}{2\sigma} - \frac{(\mu + b) (w_kx_k + \mu)}{6\sigma^3}\right) \\
\frac{\partial y}{\partial b} &\approx \sum_{k=1}^n \frac{\partial \alpha_i}{\partial b} = \Phi(\frac{\mu+b}{\sigma}) + \frac{1}{\sigma} \sum_{k=1}^n g_k
 \nonumber
\end{align}
where $g_k = \phi(\frac{\mu+b}{\sigma})\left(w_k x_k + \frac{b}{n}\right)$, and $\phi(\cdot)$ is standard  normal distribution density function. The last terms of the first two equations in \eqref{backpop} are the contribution to the gradient from the $\Phi(\frac{\mu+b}{\sigma})$ factor, which is usually small in most practical situations and thus can be safely ignored. We subsequently refer to \eqref{backpop} as the Shapley gradients.

The training process using Shapley gradients is similar to that of typical neural networks, except that \eqref{backpop} are used during back propagation stage for any layers with ReLU activation; the feed forward calculation of the neural network remains unchanged with the ReLU activation. We subsequently use the term ``Shapley Linear Unit'' (ShapLU) to refer to the training scheme of mixing Shapley gradient in the backward propagation with ReLU activation in the feed forward stage.

Even though Shapley gradient is inconsistent with the ReLU forward function, it is arguably a better choice for training neural networks; as it is more robust to descent towards the average direction of the steepest descent of all possible permutations of a neuron's inputs.

\begin{figure}

\centering

\begin{minipage}{.45\textwidth}
\caption {Attention to Exceptions\label{opinion}}
\vspace{0.5cm}

\centering

\includegraphics[width=0.95\linewidth]{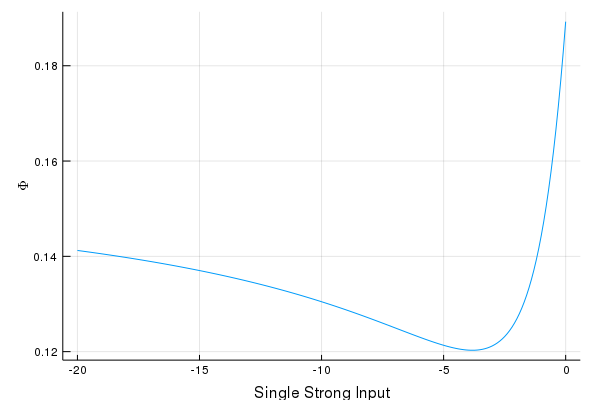}

\end{minipage}
\begin{minipage}{.45\textwidth}
\centering

\caption {Shapley Activation (SA) \label{ShapLA}}
\vspace{0.5cm}

\includegraphics[width=0.95\linewidth]{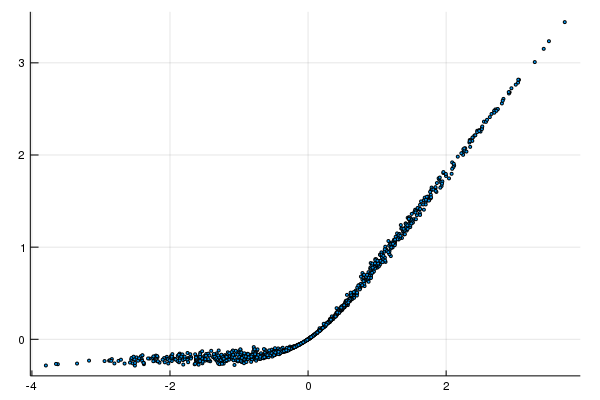}
\end{minipage}

\end{figure}

The main advantage of Shapley gradient is that it is globally continuous and never vanishes. Even when a neuron is deep in the off state with $s = \sum_{i=1}^n w_i x_i \ll 0$, significant gradient could still flow through when $\sigma$ in \eqref{app} is large. For example, when a single $w_i x_i$ signal become very strong in either positive or negative direction, the resulting increase in $\sigma$ would open the gradient flow. This is a very nice property as it is exactly the right time to update a neuron's parameters when any input signal is way out of line comparing to its peers. We call this property ``attention to exception''. Figure \ref{opinion} is a numerical illustration of this property, where we vary one $w_k x_k$ signal to a neuron and keep other inputs unchanged. The vertical axis is the $\Phi(\frac{\mu + b}{\sigma})$ factor, which controls the rate of gradient flow from the output to the input of a neuron in \eqref{app}.

\subsection{Shapley Activation}

It requires some additional efforts to implement ShapLU in most machine learning frameworks, because a customized gradient function has to be used instead of automatic differentiation. To ease the implementation of Shapley gradient, we set out to construct an activation function whose gradient matches the Shapley gradient, but maintains full consistency between the forward calculation and backward gradient. The downside of such an activation function is that it can only be an approximation to ReLU in the forward calculation.

Observe that in \eqref{app}, the cross dependency between $\alpha_i$ on $x_j, w_j$ when $i \ne j$ is only through the factor $\Phi\left(\frac{\mu  + b}{\sigma}\right)$, which is a very smooth function. The cross sensitivities $\frac{\partial \alpha_i}{\partial x_j}$ are usually small in practical settings thus can be safely ignored. Therefore, we can construct an (implied) activation function as the sum of all Shapley value $\alpha_i$s in \eqref{app}, which is a close approximation to ReLU:
\begin{equation}
\label{sa}
\max\left(\sum_{i=1}^n w_i x_i + b, 0 \right) \approx \Phi\left(\frac{\mu + b}{\sigma}\right) \left(\sum_{i=1}^n w_i x_i + b\right)
\end{equation}
Given the cross sensitivities are usually small, the gradient of \eqref{sa} closely matches the Shapley gradient in \eqref{backpop}. We subsequently call \eqref{sa} the Shapley Activation (SA), which is much easier to implement in existing machine learning frameworks via automatic differentiation.

Unlike typical activation functions that only depends on the aggregation of $s = \sum_{i=1}^n w_i x_i + b$, the activation function defined in \eqref{sa} depends on $\vec x, b, \vec w$, thus having a much more sophisticated activation profile. Typical activation functions can be plotted on a 2-D chart, but not so for \eqref{sa}. In Figure \ref{ShapLA}, we instead show a 2-D scatter plots of 1000 samples of \eqref{sa} against $s$ for $b = 0, n=5$ and $w_i x_i$ being independent uniform random variables between -1 and 1.

Figure \ref{ShapLA} bears some resemblance to leaky ReLU or noisy ReLU, however the resemblance is only superficial. Both leaky ReLU and noisy ReLU are only functions of $s$, and they can have discontinuities in gradient; while the gradient of \eqref{sa} is globally continuous, which is important for improving training convergence. \eqref{sa} is also deterministic, the apparent noise in Figure \ref{ShapLA} is from the projection of high dimensional inputs to a single scalar $s$. \eqref{sa} preserves the unique ``attention to exception'' property and allows significant gradient flow even if the neuron is deeply in the off state.

\section{Numerical Results}

\subsection{Training using Shapley gradient}
Though ShapLU and SA are close approximations to each other conceptually, they might exhibit different convergence behaviors when used in practice.

Our first test is to train a fully connected neural network to classify hand written digits using the MNIST data set \cite{mnist}. We implemented ShapLU and SA in Julia using Flux.jl \cite{flux}, which is a flexible machine learning framework that allows customized gradient function to be inconsistent from the forward calculation. In our ShapLU implementation, we neglected the last terms in the first two equations of \eqref{backpop} for simplicity and faster execution. The baseline configuration for numerical testing is a fully connected neural network of 784 input (28x28 gray scale image pixels) with two hidden layers of 100 and 50 neurons, and a output layer with 10 neurons and a softmax classifier. Both hidden layers use ReLU activation, and a cross entropy loss function is used for training.

We compared the convergence of training this neural network using 4 epochs of 1,000 unique images with a batch size of 10 and random re-ordering of batches between epochs. The entire training is repeated 10 times with different initialization to obtain the mean and standard deviation of the training accuracy. Stochastic gradient descent (SGD) \cite{robbins1951stochastic, kiefer1952stochastic, bottou2018optimization} optimizer with various learning rates(LR) were tested, as well as an adaptive ADAM optimizer \cite{kingma2014adam, bottou2018optimization} with $lr = 0.001, \beta_1 = 0.9, \beta_2 = 0.999$. In this test, the absolute classification accuracy is not the main concern, our focus is instead to compare the relative performance between ReLU, ShapLU and SA \eqref{sa} under identical settings.

\begin{figure}

\centering

\begin{minipage}{.45\textwidth}

\caption{\label{stats}MNIST Accuracy - MLP}

\centering

\vspace{.5cm}
\includegraphics[width=1\linewidth]{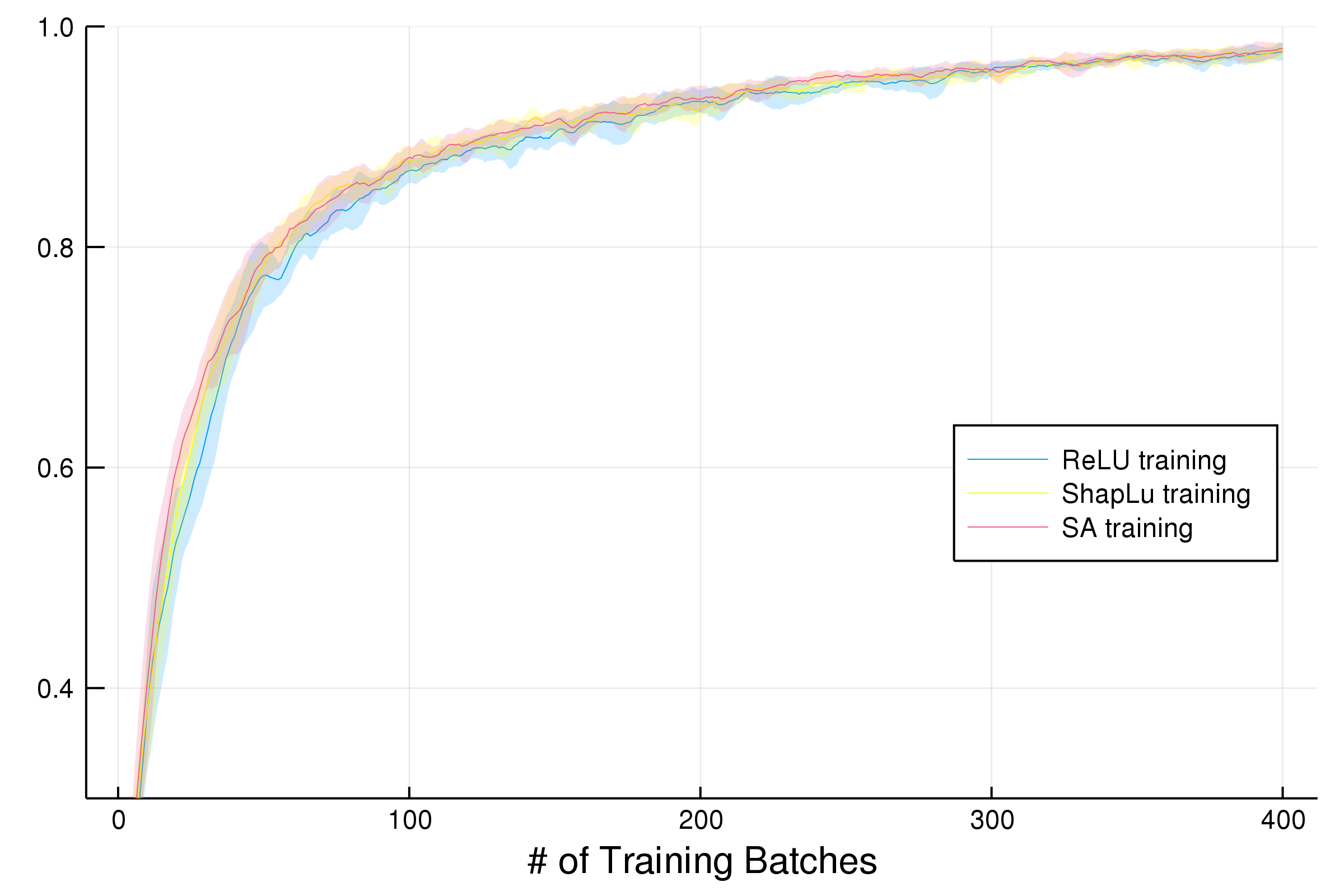}

\end{minipage}
\begin{minipage}{.45\textwidth}
\centering

\caption{CIFAR-10 Accuracy - MLP\label{cifar10}}
\centering
\vspace{.5cm}
\includegraphics[width=1\linewidth]{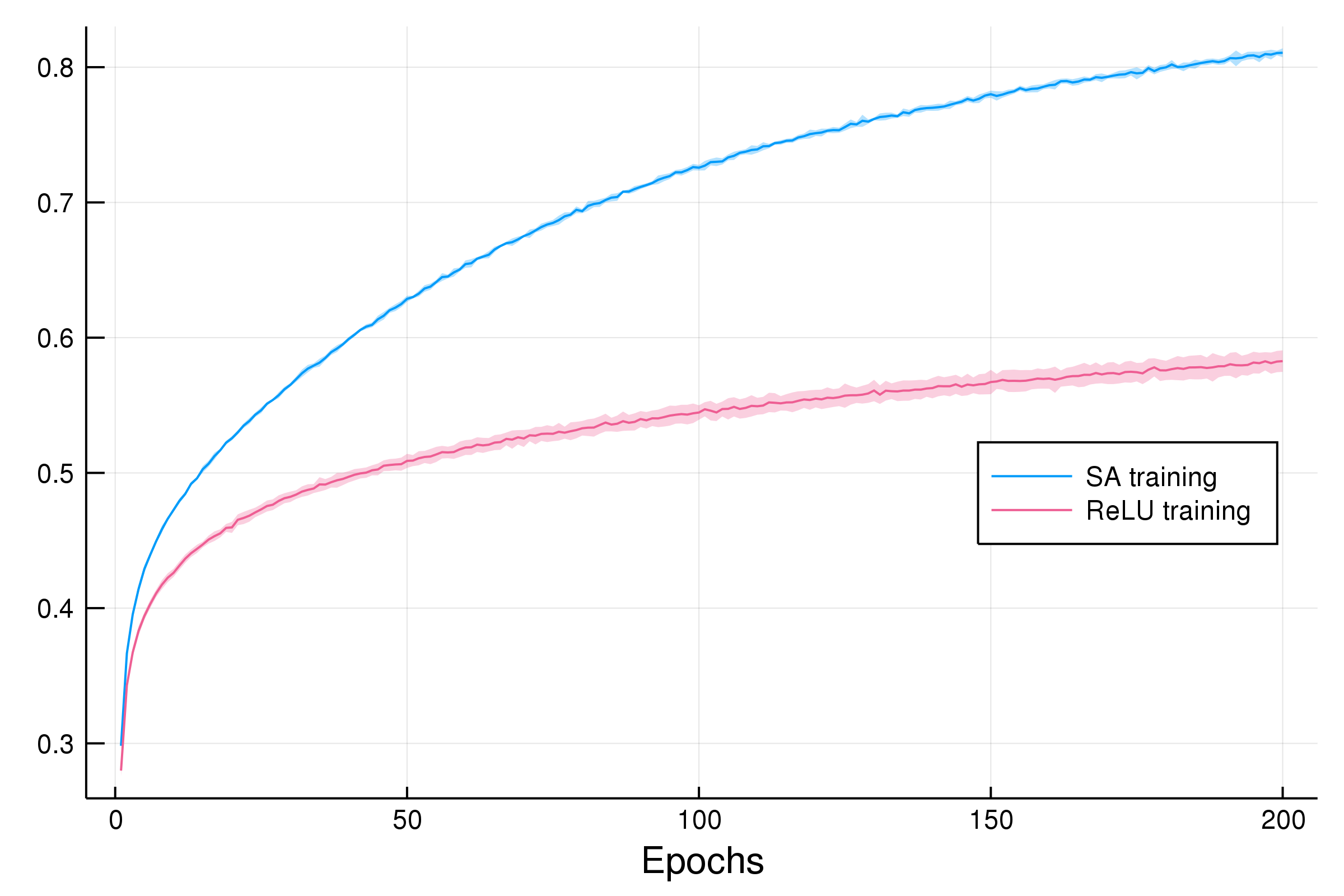}

\end{minipage}

\end{figure}

First and foremost, it is remarkable that ShapLU training actually converges. Figure \ref{stats} shows the training accuracy convergence using ADAM optimizer, where the standard deviations are shown as color shades. To our best knowledge, ShapLU is the first neural network training scheme where the back propagation uses ``inconsistent'' gradient from the forward calculation. When such consistency is broken, training usually fails. However, ShapLU outperformed ReLU in convergence using ADAM optimizer or SGD with large learning rate (LR); and they have similar convergence when smaller LR is used with SGD. This result matches our expectation that the continuous and non-vanishing Shapley gradient would lead to smoother and more stable stochastic descent. ShapLU's continuous gradient works particularly well with ADAM optimizer, resulting in visible improvements over ReLU in convergence speed during the initial phase of training, as shown in Figure \ref{stats}. The SA performed similarly to ShapLU in this MNIST test, which is not surprising as they are close approximations. The terminal validation accuracy are similar between all three methods, they all converge to about 86\% at the end of four epochs, as measured using 10,000 test MNIST images that does not include the 1,000 training images.

We then tested the SA on CIFAR-10 image classification data set \cite{krizhevsky2009learning} using Keras custom layer with TensorFlow backend. There are 50,000 training images and 10,000 test images with input shape of (32, 32, 3).
The test neural network starts with a input layer of 3072 neurons (32x32x3), then includes three hidden layers of 1024, 512 and 512 neurons, and terminate with a classification layer of 10 neurons. For each hidden layer, there is an activation function of either ReLU or SA, followed by a dropout layer with $p = 0.2$. A default glorot uniform method was used to initialize the kernel and bias. With nearly 4 million parameters, this MLP is not a trivial neural network. We trained this neural network using different optimizers with a batch size of 128, to compare the performance between ReLU and SA.

\begin{table}
\caption {CIFAR-10 Validation Accuracy - MLP \label{opt}}

\vspace{0.5cm}
\centering

\begin{tabular}{|l|l|l|}
\hline
Optimizer (lr) & Shapley Activation (SA)      & ReLU \\ \hline
SGD (0.1000)   &$0.5736 \pm 0.0053$                     & $0.5586 \pm 0.0087$                     \\ \hline
SGD (0.0500)   &$0.5703 \pm 0.0047$                     & $0.5619 \pm 0.0045$                     \\ \hline
SGD (0.0100)   &$0.5662 \pm 0.0055$                     & $0.5572 \pm 0.0084$                     \\ \hline
SGD (0.0050)   &$0.5451 \pm 0.0067$                     & $0.5460 \pm 0.0109$                     \\ \hline
Adam (0.00200) &$0.5327 \pm 0.0042$                     & $0.4559 \pm 0.0082$                     \\ \hline
Adam (0.00100) &$0.5578 \pm 0.0055$                     & $0.5102 \pm 0.0040$                     \\ \hline
Adam (0.00050) &$0.5761 \pm 0.0026$                     & $0.5389 \pm 0.0069$                     \\ \hline
Adam (0.00010) &$0.5857 \pm  0.0046$                    & $0.5692 \pm 0.0030$                     \\ \hline
RMSprop (0.00100) &$0.5369 \pm 0.0200$                  & $0.4891 \pm 0.0129$                     \\ \hline
RMSprop (0.00050) &$0.5579 \pm 0.0081$                  & $0.5209 \pm 0.0161$                     \\ \hline
RMSprop (0.00010) &$0.5743 \pm 0.0074$                  & $0.5589 \pm 0.0235$                     \\ \hline
RMSprop (0.00005) &$0.5773 \pm 0.0035$                  & $0.5718 \pm 0.0059$                     \\ \hline
\end{tabular}

\end{table}

Figure \ref{cifar10} is the training accuracy using Adam optimizer with $lr = 0.001, \beta_1 = 0.9, \beta_2 = 0.999$, where the color shadows show standard deviations computed from 20 repetitions of identical training runs. Figure \ref{cifar10} shows that SA results in a significant improvement in training accuracy, convergence and stability (i.e, smaller std dev) over those of ReLU.  Table \ref{opt} is a summary of validation accuracy at the end of training using different optimizers and learning rates \cite{robbins1951stochastic, kiefer1952stochastic, tieleman2012lecture, kingma2014adam, bottou2018optimization}, where the standard deviation is computed from 8 repetition of identical training runs. Table \ref{opt} shows that the Sharpley activation consistently outperforms ReLU in validation accuracy in almost every optimizer configuration, and many by wide margins. The SA tends to perform better with higher learning rates and it shows much less variations in validation accuracy between optimizer types and learning rates, which could be explained by its continuous and non-vanishing Shapley gradient. This example also shows that the SA works well in conjunction with dropout layers. The SA function is very efficient, we observed only a 20\% increase in CIFAR-10 training time for the same number of epochs by switching from ReLU to SA.

\begin{figure}
\caption{CIFAR-10 Accuracy - ResNet-20 \label{resnet}}

\centering

\vspace{.5cm}

\begin{minipage}{.45\textwidth}
  \includegraphics[width=1\textwidth]{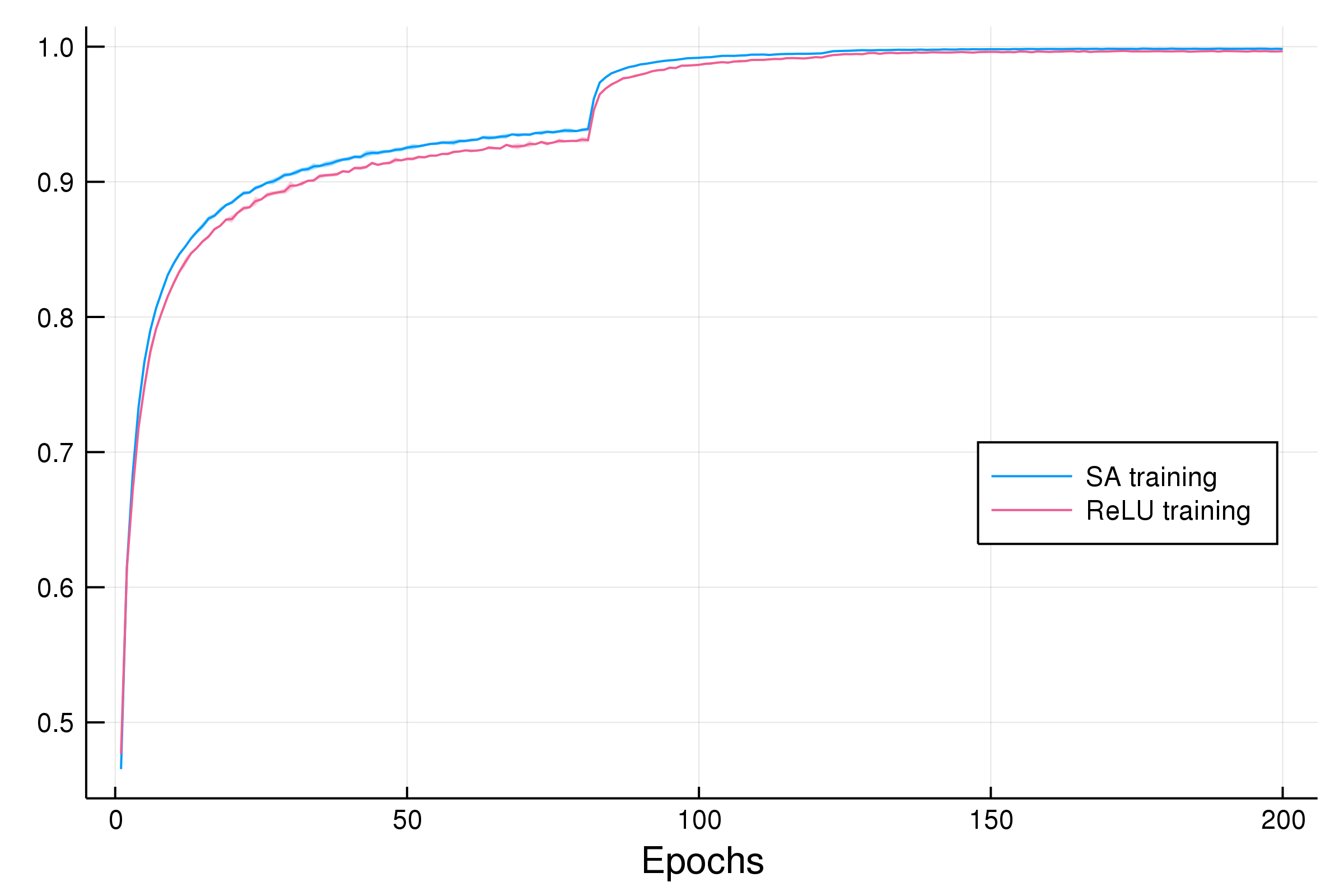}\\
\end{minipage}
\begin{minipage}{.45\textwidth}
  \includegraphics[width=1\textwidth]{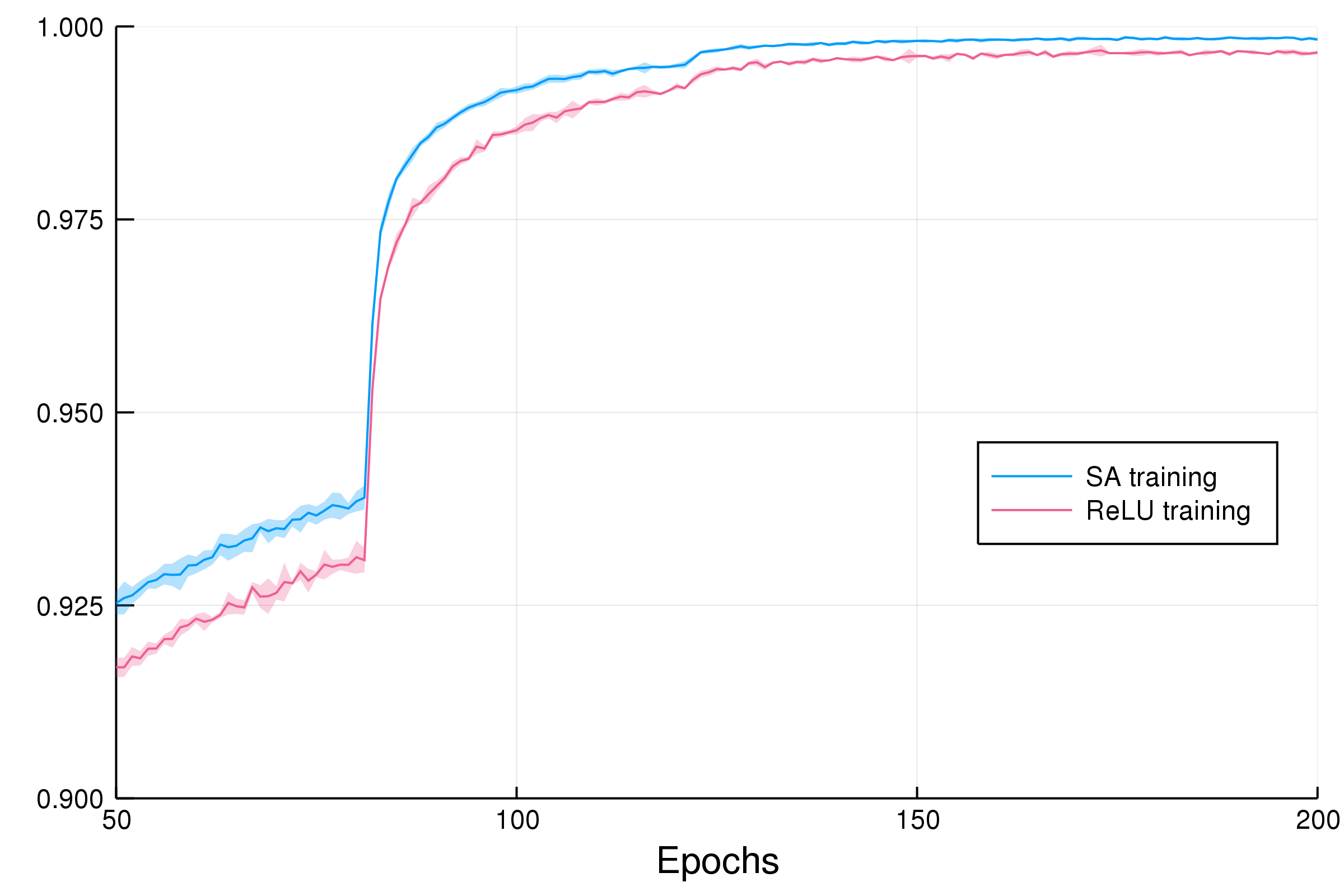}\\
\end{minipage}
\end{figure}

In addition, we implemented a convolution layer using Shapley Activation (SA) in Keras and TensorFlow and compared the convergence of ReLU and SA using ResNet-20 v2 \cite{resnet20}, which is a state-of-the-art deep neural network configuration with 20 layers of CNN and ResNet. We used the exact same configuration as \cite{gitresnet} for this test, except that we moved the batch normalization after the ReLU activation. The reason for this change is to ensure a fair comparison with SA because we chose to apply batch normalization after the SA in order not to undermine its ``attention to exception'' property. In our testing, moving batch normalization after ReLU activation results in a small improvements in training and validation accuracy compared to the original set up in \cite{gitresnet}.

Figure \ref{resnet} is the training accuracy and its standard deviation (in color shadows around the line) of ResNet-20 from 8 identical training runs using the CIFAR-10 data set, showing SA has a small but consistent and statistically significant edge in training accuracy over ReLU during the entire training process. The jumps in training accuracy at 80 and 120 epoch are due to scheduled reductions in learning rate. The right panel in Figure \ref{resnet} is a zoomed view of the training accuracy at later stage of the training. In this test, the terminal validation accuracy of SA and ReLU are both around 92.5\% and are not significantly different, presumably because the variations from the selection of validation data set is greater than any real difference in validation accuracy between the ReLU and SA. Nonetheless, because of the faster convergence and better training accuracy, ResNet-20 with SA could be trained using fewer epochs to reach a similar or higher level of training accuracy than ReLU. This example shows that even a state-of-the-art convolution neural network that is highly tuned for ReLU can further improve its training convergence and accuracy by switching from ReLU to SA, without any additional tuning. We do expect SA performance to improve further with careful tuning of its training parameters. This example also shows that SA can be used successfully in conjunction with batch normalization, and leads to overall better results.

These preliminary results validate the theory and benefits of Shapley gradients. The results from our preliminary test suggest that SA performs generally better than ReLU, and by a significant margin in certain large MLP cases. We also believe that the benefit of SA should carry over to other types of neural network architectures and applications, and more studies are needed to quantify its benefits in different network configurations.

\subsection{Interpretation using Shapley value}
\begin{figure}

\centering

\begin{minipage}{.45\textwidth}

\caption{Interpretation by Shapley Value \label{explain1}}
\vspace{.5cm}

\centering

\underline{Pixels for Accept}

\vspace{.1cm}

\includegraphics[width=.95\linewidth]{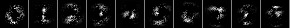}

\underline{Pixels for Rejection}

\vspace{.1cm}

\includegraphics[width=0.95\linewidth]{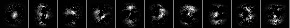}
\end{minipage}
\begin{minipage}{.45\textwidth}
\caption{Interpretation by Sensitivity \label{explain2}}
\vspace{.5cm}

\centering

\underline{Pixels for Accept}

\vspace{.1cm}

\includegraphics[width=.95\linewidth]{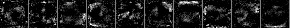}

\underline{Pixels for Reject}

\vspace{.1cm}

\includegraphics[width=0.95\linewidth]{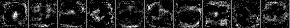}
\end{minipage}

\end{figure}

Figure \ref{explain1} is an example of interpreting a neural network's output by recursively applying LRP formula \eqref{lrp}. The neural network in this example has the same configuration as the previous MNIST MLP test, but is fully trained with MNIST data set. The Shapley values of the output softmax layer are computed using a 1000 path Monte Carlo simulation. The gray scale images are the average over 1,000 MNIST images, of input pixel's positive or negative Shapley values per unit gray scale (i.e., $\max(\frac{\alpha_k}{x_k}, 0.)$ or $-\min(\frac{\alpha_k}{x_k}, 0)$). Given the final output of this neural network is probability, the brightness in the top panel in Figure \ref{explain1} is therefore proportional to the increase in probability for a given digit if the pixel's brightness in the input image increases by 1. The bottom panel shows the same for decrease in probability. Thus bright pixels in the top panel of Figure \ref{explain1} are relevant pixels that increases the likelihood of a given image being classified to certain digit; those under bottom panel are those important pixels that decrease such likelihood.

We also show the result of a sensitivity based interpretation in Figure \ref{explain2} with identical setup for comparison. The pixel’s brightness of Figure \ref{explain2} correspond to the average magnitude of the positive (for Approve) or negative (for Reject) gradient to individual pixels of the input image.  It is evident that the Shapley value based interpretation is far superior and much more intuitive. Despite being a crude approximation, the interpretation results like Figure \ref{explain1} is sufficient to give user the much needed comfort and confidence in using neural network results for real world applications.

\section{Conclusion}

Based on an accurate analytical approximation to the Shapley value of ReLU, we established a novel and consistent theoretical framework to help address two critical problems in neural networks: interpretability and vanishing gradients. Preliminary numerical tests confirmed improvements in both areas. The same analytical approach can be applied to other activation functions than ReLU, if fast approximations to their Shapley values are known.

It is a new finding that the gradient used for stochastic descent does not have to be consistent with a neural network's forward calculation. Better training convergence and accuracy could be achieved by breaking such consistency, as shown in the example of ShapLU. Following this general direction, other inconsistent ``training gradient'' could be formulated to improve the training and/or regulate the parameterization of neural networks.

In our opinion, the Shapley value based approach is promising and future research is needed to fully understand and quantify its effects for different network architectures and applications.

\bibliographystyle{unsrtnat}
\bibliography{rwa}

\end{document}